\documentclass{article}

     \usepackage[preprint]{neurips_2022}

\usepackage[utf8]{inputenc} 
\usepackage[T1]{fontenc}    
\usepackage{hyperref}       
\usepackage{url}            
\usepackage{booktabs}       
\usepackage{amsfonts}       
\usepackage{nicefrac}       
\usepackage{microtype}      
\usepackage{xcolor}         
\usepackage{geometry}
\usepackage{amsmath}
\usepackage{graphicx}
\usepackage{subfigure}
\usepackage{mathtools}

\title{Model Monitoring and Robustness of In-Use Machine Learning Models: Quantifying Data Distribution Shifts Using Population Stability Index}

\author{
  Aria Khademi, Michael Hopka, Devesh Upadhyay\\
  Core AI/ML Research\\ Department of Research and Advanced Engineering\\
  Ford Motor Company\\
  \texttt{\{akhademi,mhopka,dupadhya\}@ford.com} \\
}

\begin{document}

\maketitle

\begin{abstract}
  Safety goes first. Meeting and maintaining industry safety standards for robustness of artificial intelligence (AI) and machine learning (ML) models require continuous monitoring for faults and performance drops. Deep learning models are widely used in industrial applications, e.g., computer vision, but the susceptibility of their performance to environment changes (e.g., noise) \emph{after deployment} on the product, are now well-known. A major challenge is detecting data distribution shifts that happen, comparing the following: {\bf (i)} development stage of AI and ML models, i.e., train/validation/test, to {\bf (ii)} deployment stage on the product (i.e., even after `testing') in the environment. We focus on a computer vision example related to autonomous driving and aim at detecting shifts that occur as a result of adding noise to images. We use the population stability index (PSI) as a measure of presence and intensity of shift and present results of our empirical experiments showing a promising potential for the PSI. We further discuss multiple aspects of model monitoring and robustness that need to be analyzed \emph{simultaneously} to achieve robustness for industry safety standards. We propose the need for and the research direction toward \emph{categorizations} of problem classes and examples where monitoring for robustness is required and present challenges and pointers for future work from a \emph{practical} perspective.
\end{abstract}

\section{Introduction}
Monitoring of AI and ML models to ensure their robustness \emph{after deployment} on customer and consumer products is \emph{safety-critical} because failure or poor performance of the products in certain circumstances can lead to catastrophic costs in high-stakes scenarios. The ML community has acknowledged the necessity and importance of model monitoring and robustness in many contexts and applications including medical practice \cite{zadorozhny2021out,finlayson2021clinician}, autonomous driving \cite{yang2021generalized,filos2020can,bogdoll2022anomaly}, etc., and this attention is growing at a fast pace. Even though some progress has been observed, major challenges of model monitoring and robustness with \emph{practical considerations} remain unresolved. Poor performance of ML models in the presence of adversaries \cite{kumar2020adversarial,goodfellow2014explaining,kireev2022effectiveness,bai2021recent,meng2022adversarial}, domain and distribution shifts \cite{mouli2021asymmetry,wortsman2022robust,taori2020measuring,miller2021accuracy,ioffe2015batch}, noises \cite{gupta2019dealing,li2021learning}, uncertainties \cite{tagasovska2019single,ovadia2019can}, among others, are still obstacles to \emph{secure and confident} deployment of ML models, especially, in cases where theoretical and practical guarantees are required.

We use an application related to autonomous driving on the publicly available LISA Traffic Light Dataset \footnote{https://www.kaggle.com/datasets/mbornoe/lisa-traffic-light-dataset} and present examples of \emph{detecting} distribution shifts in image frames from video recordings under the presence of noise. The application is motivated by the fact that environment noises are ample after deployment of AI products on vehicles and detecting them is very important to maintaining high accuracy after deployment. There are multiple ways of detecting shifts and correcting for it by measuring the distance between distributions and hypothesis testing using the KL divergence, Jensen Shannon divergence, Chi-Squared test, Wasserstein distance, importance weighting, classifier-based tests, etc., \cite{tian2021exploring,bickel2009discriminative,rabanser2019failing,sugiyama2008direct,quinonero2008dataset}. We use population stability index (PSI) \cite{yurdakul2018statistical} as the measure to detect the distribution shift in images in the presence of noise. Our empirical analyses show that PSI is a promising measure for detecting shifts, although, in its current form, it does not necessarily provide a cross-domain tool for detecting shift in general. PSI can benefit modifications including standardization for direct comparison across different applications and theoretical as well as practical studies on determining the threshold of detecting shifts using PSI. 

We further approach the problem of model monitoring and robustness in AI from an industrial perspective and assert current challenges and potential solutions with a specific focus on distributions shifts. We pin that given current results \cite{garg2022leveraging} proving the necessity of knowing some assumptions to ensure robustness, a one-size-fits-all approach to robustness does not and will not work in practice because assumptions and characteristics of data, models, and environment dynamics in each application are different. We conclude by pointing toward a tailored context-aware approach for model monitoring and robustness and providing future research directions for mitigating issues related to PSI and discussions on other aspects of robustness such as explainability. 

\section{Robustness to Environment Dynamics Post-Deployment}
Industrial applications require an ML model to not only be highly accurate, but also robust to changes after deployment. An ML model \(f\) is commonly trained, validated, and tested on \emph{apriori specified} subsets of a dataset \(D\). Even if the train/validation/test split is at random and/or is based on variants of cross-validation, it does not necessarily take into account the \emph{dynamics} of the input data, stemmed from the uncertainties of the environment, \emph{post deployment} of the model \(f\) on the product. The dynamics include (but are not limited to) the following: {\bf (i)} Data distribution shifts, where a change in distribution of data after deployment can result in a drop in performance \cite{kang2022lyapunov,rabanser2019failing,ding2021closer,mouli2021asymmetry}, {\bf (ii)} Object rotation and misplacement, which has been shown to deteriorate the performance of DNNs on unseen data \cite{alcorn2019strike,geirhos2021partial,salman2021unadversarial,engstrom2019exploring}, {\bf (iii)} Illumination and flares, that include varying lighting conditions at night or changing light angles in indoor scenarios, and lighting artifacts on images and videos that occur as a result of reflections on camera lenses or disruptive changes in environment lighting \cite{qiao2021light,wu2021train}, {\bf (iv)} Noise, which includes unseen and uncharacterized data and label noise, or sensor and operator noise \cite{wang2020training,patrini2017making,song2022learning}, {\bf (v)} Anomalies and outliers, posing a challenge to model robustness specifically on temporal non-iid data \cite{su2019robust,pang2021deep}. 

We call an ML model robust to environment dynamics post deployment if the dynamics do not significantly decrease its performance \emph{and confidence} in the performance measure (see Section \ref{perf_conf} for discussion), lower than a pre-specified safety threshold. When the distribution of the data at the development stage is different from the distribution of the data post deployment of the model, a shift has occurred, which can lead to drops in performance and confidence. Shifts manifest in different forms: 

\begin{enumerate}
    \item {\bf Concept shift:} Concept shift occurs when the joint distribution of the covariates and the labels change from the source to the target data, i.e., \(Pr_s(x,y) \neq Pr_t(x,y)\).
    \item {\bf Covariate shift:} Covariate shift refers to changes in the distribution of covariates given no distributional changes in the labels, i.e., \(Pr_s(x) \neq Pr_t(x)\) and \(Pr_s(y \mid x) = Pr_t(y \mid x)\).
    \item {\bf Label shift:} Label shift corresponds to changes in the distribution of labels while the distribution of covariates given the labels remain unchanged, i.e., \(Pr_s(y) \neq Pr_t(y)\) and \(Pr_s(x \mid y) = Pr_t(x \mid y)\).
\end{enumerate}

In this paper, we focus on data shifts, and more specifically, covariate shift, although we note that other aspects of dynamics are also important and need to be studied in future work. We scope our paper to cases where access to original and shifted data is possible. In industrial applications of autonomous driving, this includes `offline' comparison of data at development stage to those collected from the environment post deployment of the AI/ML products on vehicles.

\section{Methods and Experiments}

\subsection{Population Stability Index}
We use the population stability index (PSI) to detect data shifts. PSI measures the distance between two distributions and is computed as follows: Construct arbitrary (user-specified) bins, $\{b_i\}_{i=1}^B$ to represent the numeric values in $S$ (i.e., source). Compare the percentage of data from the source and target distributions that fall into the specified bins. If the source and target distributions are the same, the will be no difference in the number of units from the source and target distributions that fall into the bins. Let $N_s$ and $N_t$ be the total number of samples in the source and population distributions, respectively. Further, let $N_{s,b_i}$ and $N_{t,b_i}$ be the number of units from the source and target distributions that fall into bin, $i$, respectively. Then, $q_{s,b_i} = \frac{N_{s,b_i}}{N_s}$ and $q_{t,b_i} = \frac{N_{t,b_i}}{N_t}$ are the percentage of units from the source and target distributions that fall into bin, $i$. The PSI is calculated as follows:

\begin{equation}\label{psi_formula}
	PSI(S,T) = \sum_i({q_{s,b_i} - q_{t,b_i}).\ln({\frac{q_{s,b_i}}{q_{t,b_i}}})}.
\end{equation}

PSI is a symmetric version of the Kullback Leibler divergence, also discussed in \cite{yurdakul2018statistical}. Let $D_{KL}(S\|T)$ be the distance between $S$ and $T$ measured by the Kullback Leibler divergence. Then, we have:

\begin{gather*}
    D_{KL}(S\|T) + D_{KL}(T\|S) \\ 
        = \sum_x{S(x).\ln(\frac{S(x)}{T(x)})} +  \sum_x{T(x).\ln(\frac{T(x)}{S(x)})} \\
        = \sum_x{S(x).\ln(\frac{S(x)}{T(x)})} - 
        \sum_x{T(x).\ln(\frac{S(x)}{T(x)})} \\ 
        = \sum_x{[S(x) - T(x)].\ln(\frac{S(x)}{T(x)})} = PSI(S,T)
\end{gather*}

Existing work have used PSI to detect distribution shifts \cite{lin2017examining,yurdakul2018statistical,pisanie2022proposed}, but characterization of its properties in real-world scenarios, especially those on high-dimensional large scale temporal data remain fairly under-explored. We further note that the index in its original form described in Equ. \ref{psi_formula} is in $[0,\infty)$ and the numeric value of it is dependent on the number of bins. Hence, a direct comparison of the \emph{intensity of shift} measured by PSI in different applications is not trivial.

\subsection{Experiments on Real-World Data}\label{experiments}
We focus on autonomous driving and present analyses on the publicly available LISA Traffic Light dataset. The dataset provides day-time and night-time sequences of frames from traffic videos captured in San Diego, CA. We chose day-time and night-time images (i.e. frames) and added Gaussian, speckle (where adding noise to image includes following the formula, image + n * image, where n is Gaussian noise with some mean and variance), as well as salt and pepper (S\&P) noises to them in a controlled fashion. We ran two types of experiments all with default Python implementation settings unless noted otherwise. \footnote{In all experiments, we used an implementation of PSI available at https://github.com/mwburke/population-stability-index. We further used a standard Python implementation of noise at https://scikit-image.org/docs/dev/api/skimage.util.html\#random-noise.} Note that PSI captures the difference between distributions of numeric values. Since the images of the LISA dataset are colored, in all experiments, we separated each input image to its three RGB channels and performed our experiments on each channel separately. For gray-scale images, one can perform these experiments on the pixel values of the image.

{\bf (i) Multiple noise intensity levels on single image experiment}: In this experiment, to analyze PSI on various levels of noise, we chose one day-time and one night-time image (shown in Appendix \ref{single_img_experiment_image_examples}, Figures \ref{fig:day_img} and \ref{fig:night_img}). Then, for Gaussian and speckle noises, we kept a zero mean and changed the variance in $\{0,0.1, \ldots, 0.9,1\}$. For salt and pepper noise, we note the following: Two important factors that determine the intensity of salt and pepper noise are i) `proportion': proportion of salt vs. pepper, and ii) `amount': proportion of pixels to be randomly chosen and replaced with noise among all image pixels. For example, proportion $0.1$ means that the proportion of salt vs. pepper is $0.1$ and amount $0.1$ means that 10\% of image pixels are replaced with noise. In our experiments, we changed the values of both proportion and amount in $\{0,0.1, \ldots, 0.9,1\}$ and computed PSI between the original image and the image with added noise. We show the results of these experiments in Figs. \ref{fig:psi_lines} and \ref{fig:my_sp_3d}.

{\bf (ii) Multiple images with a single noise intensity experiment}: In this experiment, to analyze the distribution of PSI on various images, we added noise to $10,954$ day-time images (days 1 and 2 combined) and $11,527$ night-time images (nights 1 and 2 combined), and computed the PSI for each image comparing its original form and the one with added noise. For Gaussian and speckle noises, we used mean $0$ and variance $0.1$ and for salt and pepper noise, half of the image pixels were randomly chosen for adding noise while other settings were kept as default. We show the results of this experiment in the box plots of Fig. \ref{fig:psi_box_plots}. 

\subsection{Results}
\begin{figure}
    \centering
    \subfigure[]{\includegraphics[width=0.45\textwidth]{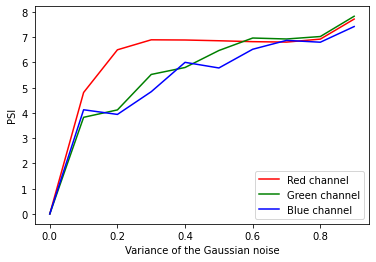}} 
    \subfigure[]{\includegraphics[width=0.45\textwidth]{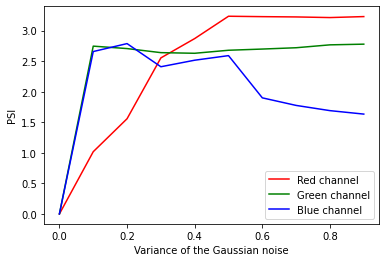}}
    \subfigure[]{\includegraphics[width=0.45\textwidth]{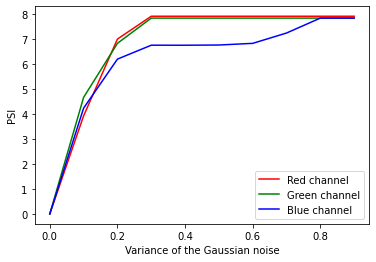}}
    \subfigure[]{\includegraphics[width=0.45\textwidth]{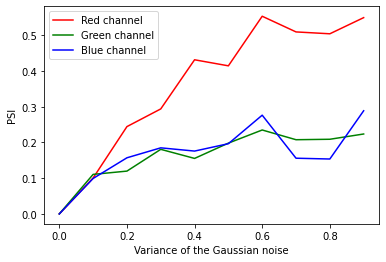}}
    \caption{PSI values in the `multiple noise intensity on single image' experiment. X axes are noise levels, i.e., variance in Gaussian and speckle noise. Y axes are PSI values. Top and bottom rows correspond to day-time and night-time image, respectively. (a) Gaussian noise during the day. (b) Speckle noise during the day (the X axis in the image is the variance of the Gaussian noise that generates speckle noise). (c) Gaussian noise at night. (d) Speckle noise at night (the X axis in the image is the variance of the Gaussian noise that generates speckle noise).}
    \label{fig:psi_lines}
\end{figure}

\begin{figure}
    \centering
    \subfigure[]{\includegraphics[width=0.45\textwidth]{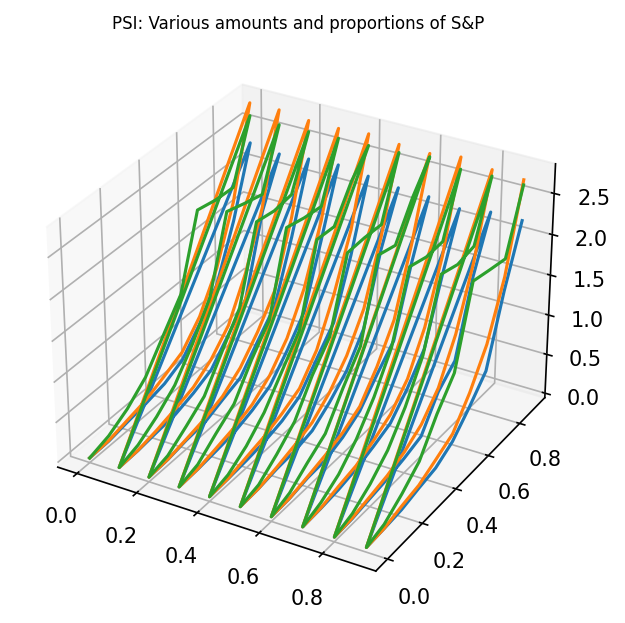}} 
    \subfigure[]{\includegraphics[width=0.45\textwidth]{images/sp_noise_day_varying_prop_amount_day.png}}
    \caption{PSI values with varying levels of S\&P noise intensity on an image during the day (left) and an image at night (right). Levels of amount and proportion, defined in Section \ref{experiments} range in $\{0,0.1, \ldots, 0.9,1\}$ on the X and Y axes and PSI values are on the Z axis illustrated for all RGB channels.}
    \label{fig:my_sp_3d}
\end{figure}

We show the results of our experiments using multiple noise intensity levels and noise types on a single image, i.e., experiment (i) described above, in Figs. \ref{fig:psi_lines} and \ref{fig:my_sp_3d}. We observe that PSI mostly increases with increase of variance in Gaussian and speckle noises, and with increases of amount and proportion in S\&P noise, in all three RGB image channels both during the day and at night. We note that the scalar value of PSI increases above 1 even with small values of variance in Gaussian noise both during the day and at night. There is a similar trend during the day for speckle noise. This shows the sensitivity of PSI to increases in Gaussian and speckle noise which can be helpful in detecting even small amounts of shifts. A similar trend of sensitivity, although not with the exact numbers, is observed for S\&P noise in Fig. \ref{fig:my_sp_3d}. The sensitivity of PSI to noise intensity can be helpful in applications when (even small) errors can lead to catastrophic costs. 

\begin{figure}
    \centering
    \subfigure[]{\includegraphics[width=0.3\textwidth]{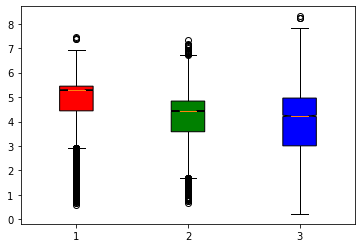}} 
    \subfigure[]{\includegraphics[width=0.3\textwidth]{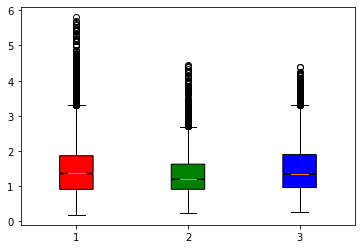}}
    \subfigure[]{\includegraphics[width=0.3\textwidth]{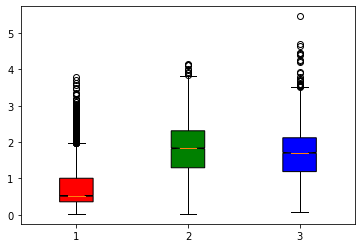}}
    \subfigure[]{\includegraphics[width=0.3\textwidth]{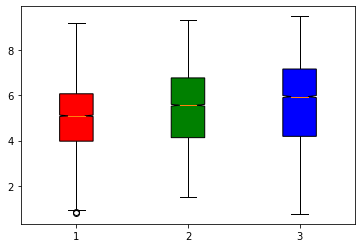}}
    \subfigure[]{\includegraphics[width=0.3\textwidth]{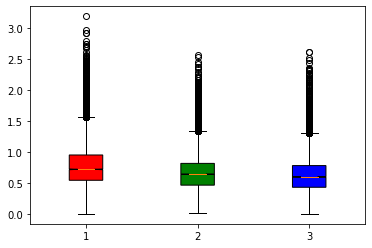}}
    \subfigure[]{\includegraphics[width=0.3\textwidth]{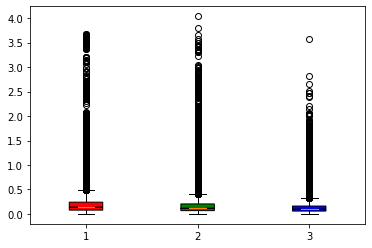}}
    \caption{PSI values for each RGB channel of the `multiple images with a single noise intensity' experiment. The X and Y axes represent RGB channels and PSI values, respectively. Top and bottom rows correspond to day-time and night-time image, respectively. (a) Gaussian noise during the day. (b) Speckle noise during the day (the X axis in the image is the variance of the Gaussian noise that generates speckle noise). (c) S\&P noise during the day. (d) Gaussian noise at night. (e) Speckle noise at night (the X axis in the image is the variance of the Gaussian noise that generates speckle noise). (f) S\&P noise at night.}
    \label{fig:psi_box_plots}
\end{figure}

Fig. \ref{fig:psi_box_plots} shows the results of our experiments with multiple images using a single noise intensity level from different noise types, i.e., experiment (ii) described in the previous sections. The distribution of PSI across multiple images shows that PSI is indeed able to capture distribution shifts. PSI shows high and moderate sensitivity to Gaussian as well as salt and pepper noises in various environment lighting, i.e., day and night. PSI is further moderately sensitive to speckle noise during the day and shows lower sensitivity at night.

To our knowledge, there is currently no established and reliable mechanism for determining PSI thresholds that indicate distribution shifts as a result of noise. In the analyses above, we used high, moderate, or low sensitivity of PSI for our interpretations in an ad-hoc manner. Determining the sensitivity threshold for PSI with respect to different noise types, noise intensities, environment lighting, for direct comparison in various applications remains an active area of research. On the other side, the observations in our experiments point to the advantage of PSI in capturing the difference between distributions resulted from different noise types, noise intensities, and various environment lighting (day and night). In interpreting the results, we note the dependency of PSI to the number of bins and the fact that it does not capture the \emph{locality} of changes in images. We discuss the caveats in more detail in the Discussion section.

\section{Discussion and Conclusion}

\subsection{PSI}
We experimented on real-world computer vision data related to autonomous driving and aimed at capturing distribution shifts that occur as a result of noise using PSI. We observed the advantages and weaknesses of PSI and note the following important points: {\bf (i)} PSI is dependent on the number of bins that characterize the distribution of data in Equ. \ref{psi_formula}. Currently, the sensitivity of the results of PSI and their \emph{implications to the application} are under-explored. {\bf (ii)} PSI computes the change in distributions in terms of the percentages of the data that fall into the pre-specified bins. This is limited by the fact that the \emph{locality} of changes is not captured in the binning process, or any other means of representing distributions that do not show \emph{where} the change in the input image happened. Determining the locality can be helpful in diagnosing robustness issues related to convolutional neural network applications in computer vision. For example, if re-training of the network is required for robustness, one can capture the locality of change, and only re-train the layers that represent the locality. This can save computation time.

\subsection{Performance and Confidence Metrics} \label{perf_conf}
Industrial and real-world practice of machine learning requires two measures for assessment of efficacy and reliability: 

\begin{itemize}
    \item Performance metric, e.g., accuracy, F-measure, AUC, etc., to assess correct predictions, and
    \item Confidence measure to assess our confidence on the measured performance metric.
\end{itemize}

The two measures of performance \emph{and} confidence must be obtained and presented together. For example, if a machine learning model has classified an input image as a `cat', one must say that the model has 99\% accuracy (on for instance a collection of unseen test data at the development stage) \emph{and} that on this specific new image, we are 99\% \emph{confident} that the image is indeed a cat. Note that the confidence measure indicates how strongly the machine learning model \emph{believes} that the prediction is indeed true; a concept that moves beyond merely reporting performance measures. This concept is important because in real-world practice, \emph{high confidence on wrong predictions} can lead to unbearable costs. These costs cannot be \emph{prevented} by merely reporting the rates of false positives, false negatives, and alike since the industrial product has to take an immediate action in the case of unseen samples in practice. Recent works have studied confidence measures \cite{garg2022leveraging,guillory2021predicting}, but this idea is yet unsettled in the machine learning community from the academic and industrial sides, especially from a practical standpoint.

Absence of ground truth is a major challenge in reporting confidence post deployment of the product. Recent work have touched on the concepts of `accuracy-on-the-line' \cite{miller2021accuracy} and `agreement-on-the-line' \cite{baek2022agreement} for predicting out-of-distribution performance without access to labeled test data. While the lines of work are interesting, they are limited in the following aspects. Accuracy-on-the-line does not universally hold and there is no theoretical guarantee for it on edge cases, corner cases, adversarial examples, etc. Agreement-on-the-line, an empirical look that only focuses on neural networks, takes \emph{agreement} of a pair of networks instead of their accuracy and so, suffers from the following possible scenario: Two neural networks \emph{can both agree on the wrong prediction}. In this case, we will predict wrongly with high confidence. Increasing the number of neural networks with different architectures can assist in achieving better confidence, but providing guarantees remain unsettled. From an industrial standpoint, if guarantees in some cases cannot be achieved to meet industry safety standards, procedures must be incorporated to prevent costs.

\subsection{Characterizing Real-World Challenges}
How much do theoretical developments and empirical results on becnhmark data \emph{indeed} capture the challenges of real-world applications? With growing progress in benchmarking datasets \cite{koh2021wilds,hendrycks2018benchmarking}, specifying and formulating the \emph{characteristics} of shifts in high-stakes real-world applications including medical practice and autonomous driving, where a single fault can lead to catastrophic costs, remains a path for future research. This characterization includes precise specification of shifts (i.e., in mean, variance, \emph{type} of the distribution, etc.) and the \emph{real-world situations} in which those shifts occur. Take autonomous driving as an example: Research must specify in detail the shift(s) that can occur in a crowded intersection in the presence of heavy fog during the day compared to the shift(s) that can occur on the road in the presence of sleet at night with flares in the input images, both on non-iid data over time. Research must further specify the methods that can determine the type of shift(s) in high-dimensions, \emph{and} the specific type of methodologies (e.g., re-training, batch-normalization, robust training, etc.) tailored to mitigate the issue(s) related to performance degradation and confidence score reports. Note that from an industrial perspective, the characterization is required for massive high-dimensional non-iid temporal data with no ground truth \emph{after deployment} on the product. Cases where guarantees cannot be made with high-confidence need to be explicitly stated to prevent casualties. 

\subsection{Relationship Between Performance and Data Shift}
If the performance of a deployed machine learning model decreases because of distribution shifts, noises, adversaries, etc., post deployment on the product and considering whether or not the confidence metric (positively or negatively) correlates with the decreased performance, the main challenge is answering the following question: What is the \emph{precise relationship}, between performance (and confidence) change (i.e., exactly how much, if any) and the amount of change in the data, e.g., distribution shift? For instance, if the object in an unseen image has been rotated \(\delta\) degrees, and if the performance of the machine learning model has dropped by \(\gamma\) as a result of the rotation, we then need a mapping function \(\gamma = M(\delta)\), where \(M(\cdot)\) can be a deterministic or stochastic function that needs to be estimated (i.e., learned) or uniquely identified. A similar question is of concern about the relationship between changes in data and confidence metrics.

The interpretation of this relationship is context dependent. Imagine a self-driving car misreads the word `stop' in a stop sign with the word `speed limit 45'. On the other hand, imagine the driver assisting technology in parallel park miscalculates the distance between the bumper and the curb due to fog, so the vehicle may hit the curb. The former scenario seems more costly than the second scenario because it is more likely to lead to a crash with a pedestrian. The precise relationship between the shift and the error can prevent casualties in various levels of granularity. 

\subsection{Model Monitoring and Explainability}
One aspect of model monitoring and robustness is detecting and correcting for faults after deployment. Explainability of ML models is an important factor that aids fault detection. The more explainable the ML models are, the more efficiently and reliably one can detect root causes of faults in real-time and adjust for them. Deep learning models are one of the main ML tools used in practice. However, the more complex deep learning models are, the harder it gets to detect the root cause of faults and failures in real-time. The complexity of neural networks has been proven in general to be necessary in order to achieve accuracy \cite{bubeck2021universal} and such complexity is an obstacle to model `explainability.' Given neural networks with millions, if not billions, of parameters (necessary for accuracy) in industrial practice, explaining the behavior and outcomes of the neural networks, especially explaining the reason for their vulnerability to and poor performance on corner cases and out-of-distribution test cases post deployment, is not trivial. Therefore, we conjecture that given the proof in \cite{bubeck2021universal}, mitigating the trade-off between complexity and model explainability, i.e., making simpler models to make them explainable, is not possible because maintaining high accuracy cannot be overlooked. In some cases, the internal structure of the complex models is not necessarily possible in general (at least in real-time) and hence, explaining the models \emph{independent} of their internal structure or complexity, e.g., independent of the number of tuned parameters is preferred. Recent methods have been proposed \cite{khademi2020causal} for explanation of black box models without accessing the internal structure of the model using causal inference. Such approaches to model explanation benefit from their independence of the structure of the black box model and show a fruitful path for future research. 

\subsection{Toward a Context-Aware Tailored Approach to Robustness}
We posit that \emph{a one-size-fits-all approach to achieving robustness does not and will not work} in high-stakes real-world and industrial applications. Generalization and robustness to out-of-distribution data will be achieved through characterization and parametrization of \emph{problem classes} and proposing solutions according to the context and application. We therefore assert {\bf the need for characterization of problem classes}: First, some of the environment and data uncertainties post deployment of the machine learning model(s) on the product are essentially \emph{unpredictable}. For example, characterization of all possible noises and data distributions under various storm weather conditions for computer vision applications in autonomous vehicles is not possible. Related work of \cite{thams2022evaluating} has discussed this by proposing to construct the set of `realistic' future test distributions using parametrization of the distributions that are conceivable in practice. Also, \cite{kang2022lyapunov} proposes to construct a feasibility space using Lyapunov density models where it is possible to guarantee the performance of machine learning models, as opposed to infeasible spaces where presenting such out-of-distribution performance guarantees are not possible. Second, robustness solutions that require heavy software and hardware involvement are not necessarily applicable to some scenarios with software and hardware constraints. We further note that ground truth is not available in real-world applications after deployment, in which case, robustness solutions have to be developed under simulated settings with relevant examples. The simulated nature of the development phase is an obstacle to generalizability of the potentially proposed robustness methods. Given the above, we propose that research should characterize ML problem classes as well as their contexts and then, provide robustness solutions that are tailored to the characterization and context.

\newpage


\bibliographystyle{IEEEtran}


\newpage

\appendix

\section{Image and noise examples} \label{single_img_experiment_image_examples}
In this section, we present the images that we used in the `multiple noise intensity levels on single image experiment' both during the day and at night in the Figures \ref{fig:day_img} and \ref{fig:night_img} below. Also, note that we aggregated all of the images in day sequence 1 and day sequence 2 for day-time images as well as all of the images in the night sequence 1 and night sequence 2 for night-time image of the `multiple images with a single noise intensity experiment'.

\begin{figure}[h]
    \centering
    \subfigure[]{\includegraphics[width=\textwidth]{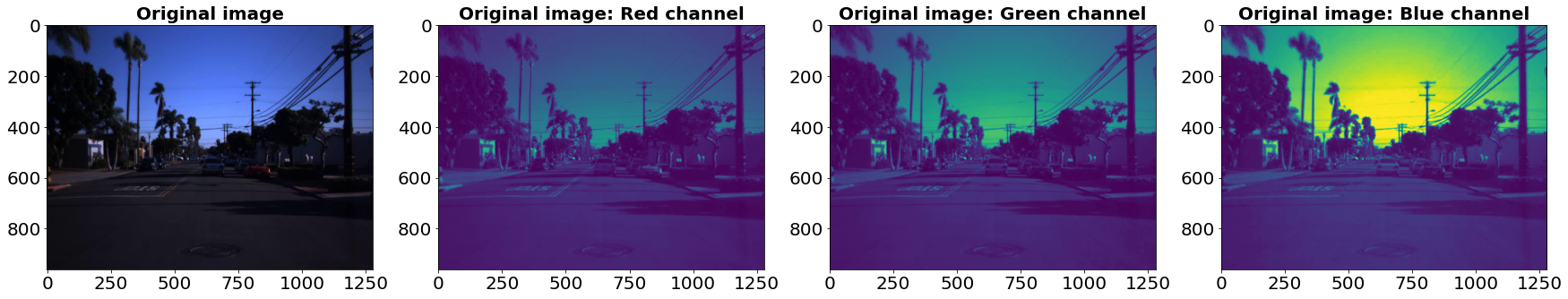}} 
    \subfigure[]{\includegraphics[width=\textwidth]{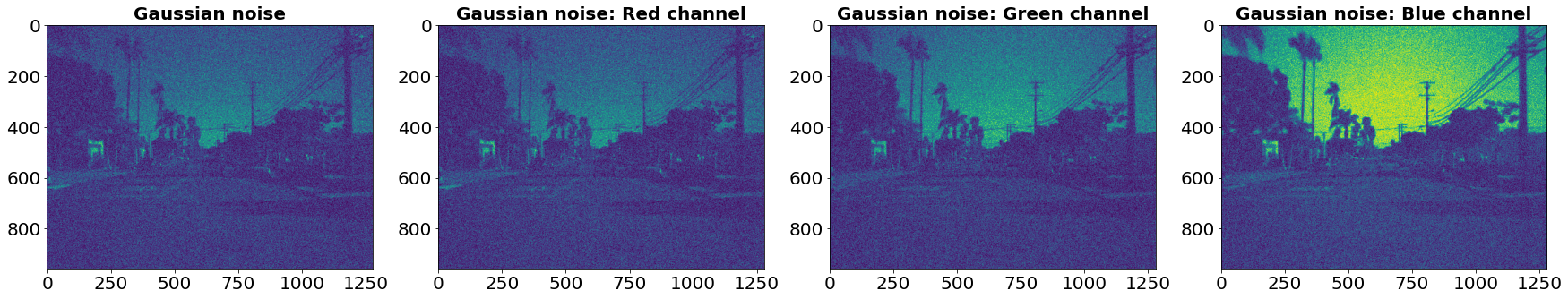}} 
    \subfigure[]{\includegraphics[width=\textwidth]{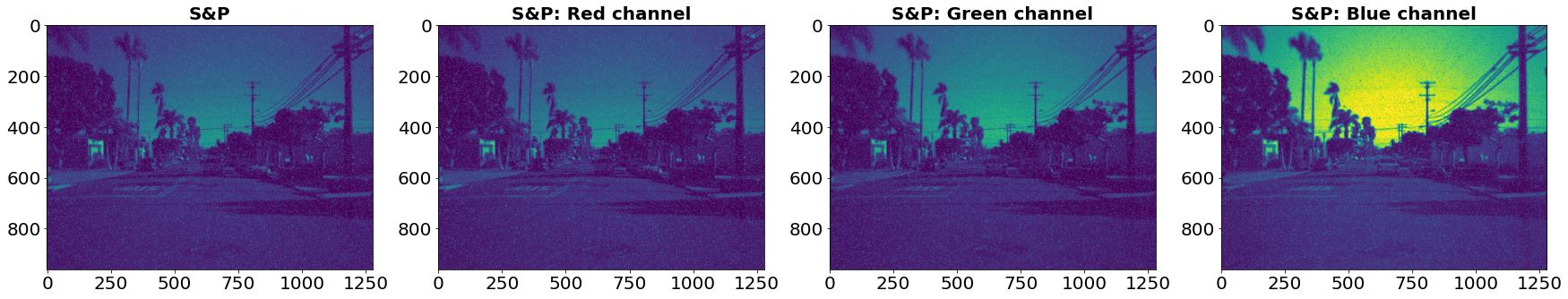}}
    \caption{The day-time image used in the `multiple noise intensity levels on single image experiment'. (a) Original image and its corresponding RGB channels. (b) The image with added Gaussian noise and its corresponding RGB channels. (c) The image with added S\&P noise and its corresponding RGB channels.}
    \label{fig:day_img}
\end{figure}

\begin{figure}
    \centering
    \subfigure[]{\includegraphics[width=\textwidth]{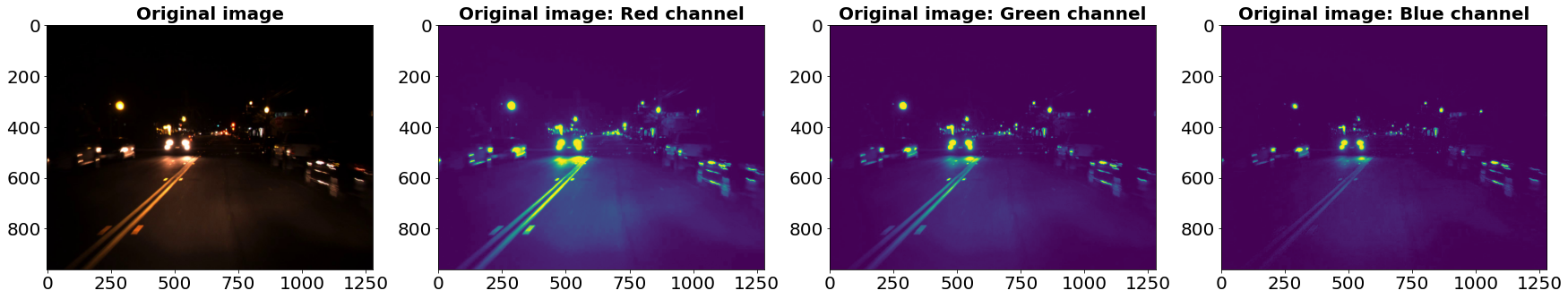}} 
    \subfigure[]{\includegraphics[width=\textwidth]{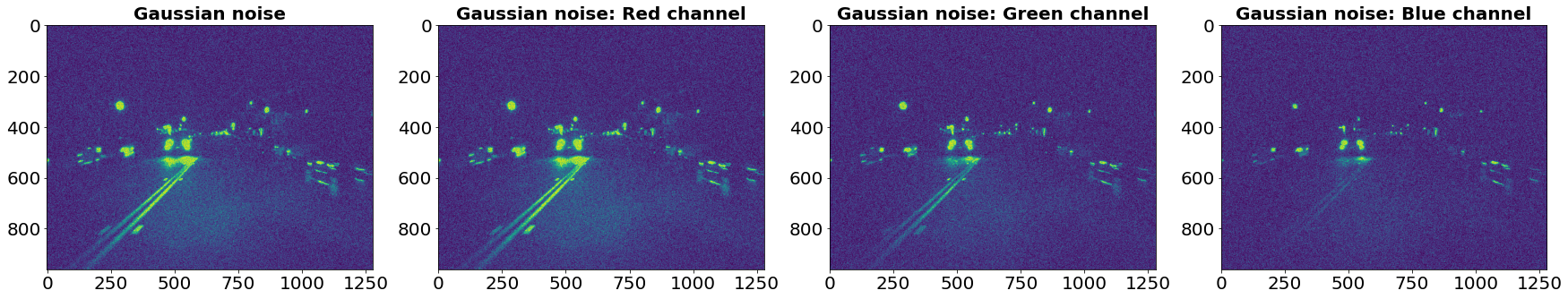}} 
    \subfigure[]{\includegraphics[width=\textwidth]{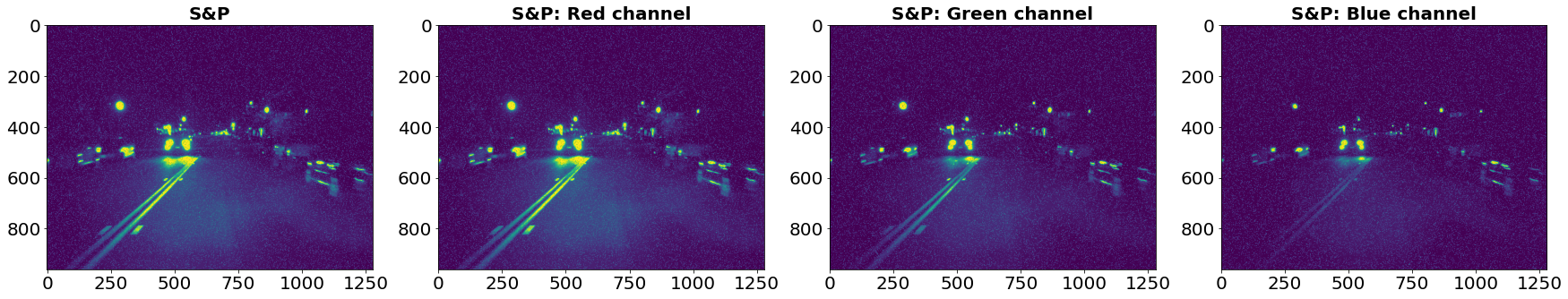}}
    \caption{The night-time image used in the `multiple noise intensity levels on single image experiment'. (a) Original image and its corresponding RGB channels. (b) The image with added Gaussian noise and its corresponding RGB channels. (c) The image with added S\&P noise and its corresponding RGB channels.}
    \label{fig:night_img}
\end{figure}

\end{document}